\documentclass[10pt,a4paper]{seedance-preprint}
\usepackage{preprint}
\usepackage{longtable}
\institutionlogos{
\raisebox{-0.5\height}{\includegraphics[height=8mm]{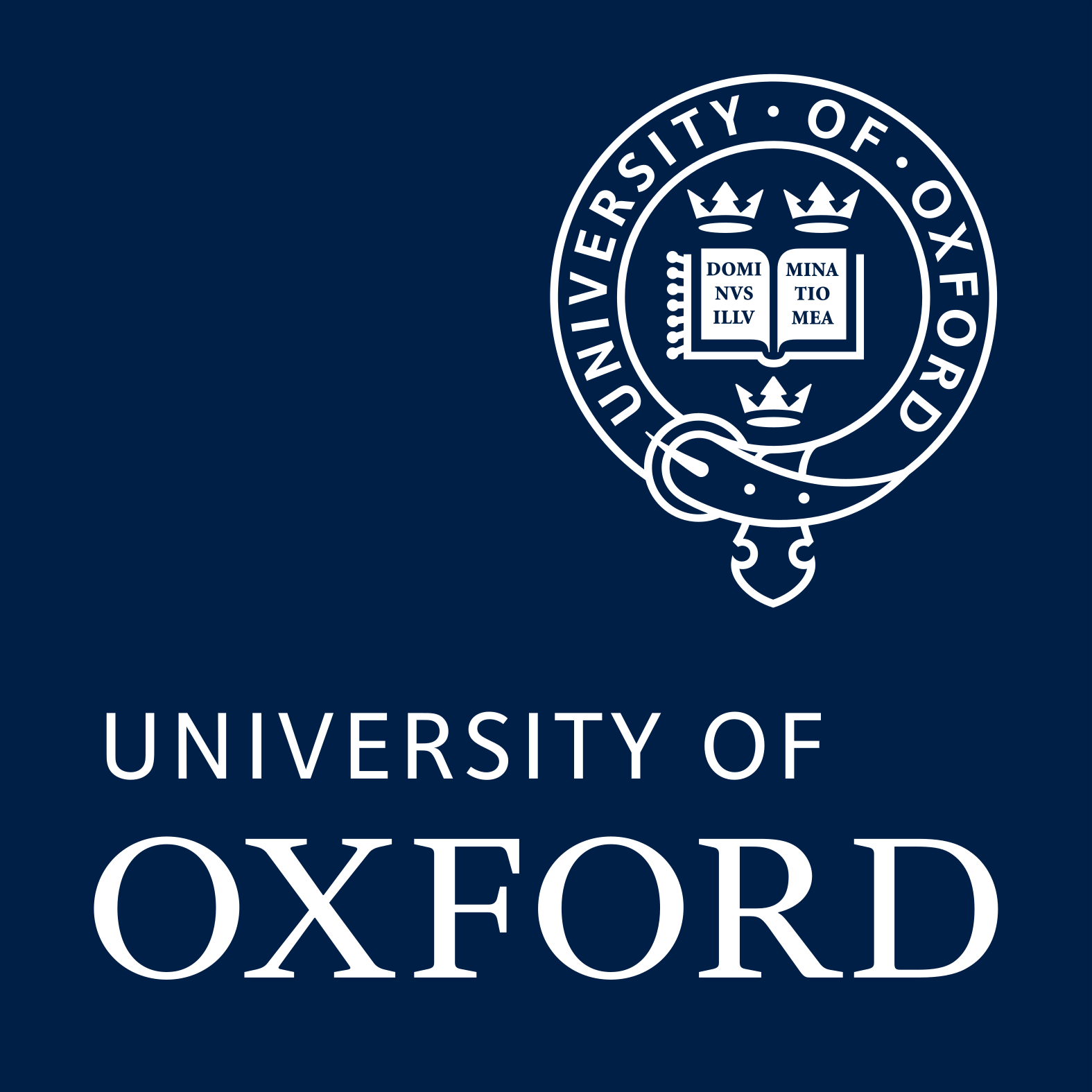}}%
\hspace{4mm}%
\raisebox{-0.5\height}{\includegraphics[width=27mm]{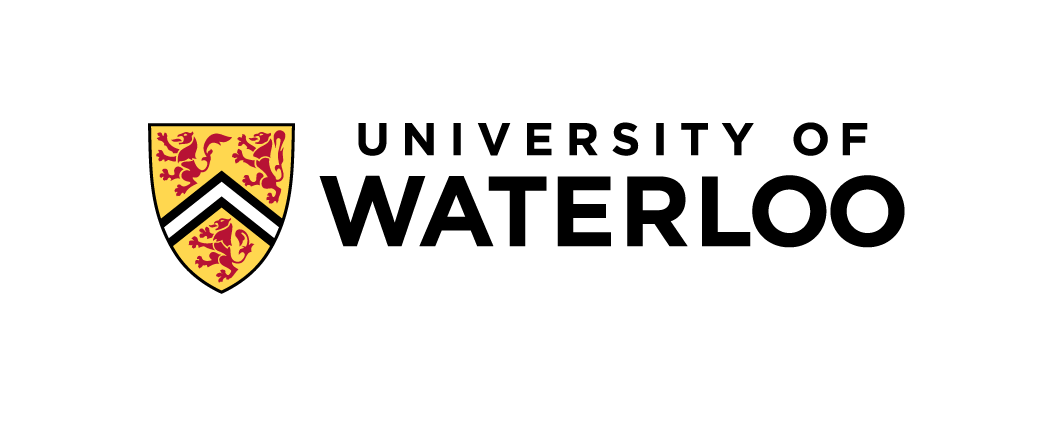}}%
\hspace{4mm}%
\raisebox{-0.5\height}{\includegraphics[width=17mm]{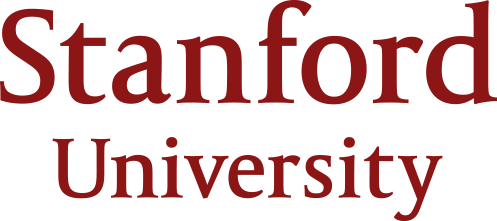}}%
}
\ifdefined\ReviewCopy\reviewcopytrue\fi


\begin{document}

\title[AI Theorist for quantum materials]{The AI Theorist reveals excitonic structure\texorpdfstring{\\[3pt]}{ }in \texorpdfstring{$\alpha$-RuCl$_3$}{α-RuCl3}}

\author[1,*,\dagger]{Hongjian Zhou}
\author[2,*]{Xianfan Nie}
\author[3]{Sean Wu}
\author[2]{Tarun Patel}
\author[4]{Jinge Wu}
\author[1]{Andrew Liu}
\author[2]{Adam Wei Tsen}
\author[1]{David A. Clifton}
\affiliation[1]{University of Oxford, Oxford, United Kingdom}
\affiliation[2]{University of Waterloo, Waterloo, Canada}
\affiliation[3]{Stanford University, Stanford, USA}
\affiliation[4]{University College London, London, United Kingdom}

\abstract{
Advances in experimental instrumentation and automation generate increasingly rich datasets, but turning experimental observations into microscopic understanding remains a bottleneck in scientific discovery. To accelerate this process, we introduce AI Theorist, a system of artificial intelligence (AI) agents for autonomous discovery of physical models through hypothesis generation, first-principles calculations and evidence-driven refinement. We apply the framework to $\alpha$-RuCl$_3$, a leading candidate material for realizing a Kitaev quantum spin liquid, to investigate its electronic structure through optical spectra. AI Theorist develops a new interpretation of the optical and photocurrent observations, identifying distinct excitonic states with contrasting optical selection rules and real-space distributions. To our knowledge, this is the first demonstration of an AI system autonomously developing a physical model to explain previously unpublished experimental observations in a quantum material, utilizing first-principles electronic-structure and many-body calculations. Our results establish a route to autonomous theoretical discovery in materials science, in which AI agents use first-principles calculations to turn experimental observations into physical models and testable predictions.
}


\maketitle
\begingroup
\renewcommand{\thefootnote}{\fnsymbol{footnote}}
\footnotetext[1]{These authors contributed equally to this work.\quad\textsuperscript{\textdagger}Corresponding author: Hongjian Zhou.}
\endgroup
\ifreviewcopy\linenumbers\fi
\begin{figure}[!p]
\centering
\includegraphics[width=0.96\textwidth]{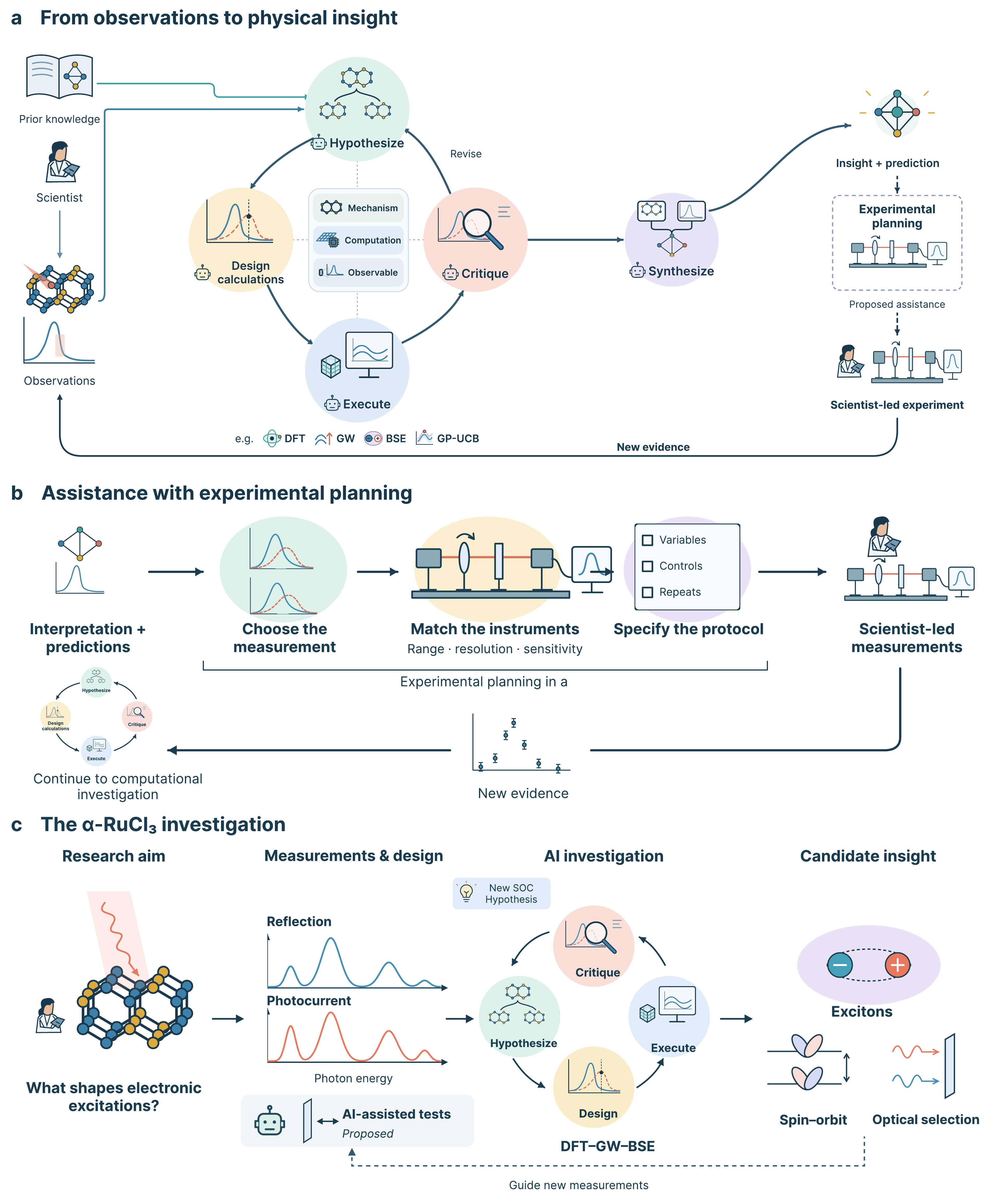}
\caption{\textbf{Architecture and workflow of AI Theorist.} \textbf{a,} Human scientists set the overall research direction, plan experiments with AI Theorist and perform the measurements. Drawing on the resulting observations and prior knowledge, AI Theorist autonomously develops physical models through iterative hypothesis generation, calculation design, execution and critique. Robot badges denote agent roles. \textbf{b,} AI Theorist assists with measurement, instrument and protocol selection. Scientists finalize the protocols and perform measurements, providing new evidence for the computational investigation. \textbf{c,} The $\alpha$-RuCl$_3$ investigation proceeds from the broad aim of understanding electronic excitations in a correlated quantum material, through AI-assisted experimental design and measurements, to autonomous model development. Reflection and photocurrent spectra serve as complementary probes of electronic excitations. AI Theorist formulates and tests hypotheses, including the role of spin--orbit coupling, using DFT--$GW$--Bethe--Salpeter calculations. Candidate interpretations address excitonic character, spin--orbit effects and optical selection rules.}\label{fig:architecture}
\end{figure}

Turning experimental observations into microscopic understanding remains a bottleneck in scientific discovery and requires several kinds of scientific expertise. A successful discovery usually involves researching literature, interpreting measurements, constructing theoretical models, and performing calculations. Connecting these contributions requires decisions about the appropriate level of analysis, approximations that can be justified, and quantities that should be compared with experiments.

Large language models (LLMs) offer a basis for systems that autonomously carry out these tasks. Their ability to work with scientific language and code connects literature analysis, computational implementation and interpretation of results. Multistep reasoning and tool use allow agents to plan investigations, execute calculations and revise their next steps using external evidence~\cite{wei2022chain,yao2023react}. LLM-based research systems have generated mathematical constructions, developed nanobody-design workflows and proposed biomedical hypotheses~\cite{romeraparedes2024mathematical,swanson2025virtual,gottweis2026coscientist}. These results motivate autonomous investigations that connect observations to explanatory physical models.

Previous work establishes that AI systems can infer scientific relations, conduct research workflows and generate experimentally testable explanations. Equation-discovery methods recover mathematical relations from data, sometimes incorporating existing theory~\cite{schmidt2009laws,udrescu2020aifeynman,cornelio2023descartes}. The AI Scientist automates computational research in machine learning~\cite{lu2026automation}, while Virtual Lab develops nanobody-design workflows through collaboration between specialist agents and human researchers~\cite{swanson2025virtual}. Virtual Biotech coordinates specialist agents to integrate biomedical and clinical evidence for therapeutic discovery and development~\cite{zhang2026virtualbiotech}. Co-Scientist and Robin extend these capabilities to biological hypotheses and mechanisms assessed through experiments~\cite{gottweis2026coscientist,ghareeb2026robin}. These studies leave open whether an AI system can autonomously develop and computationally test a microscopic model that explains new experimental observations. This requires selecting physical interactions and approximations from experimental evidence, testing competing explanations through their calculated observable consequences, and revising the model against evidence. The theory must yield physical insight beyond the observations themselves and generate testable predictions that distinguish competing explanations and guide further experiments. These physical models have traditionally been hand designed by human theorists; automating this progression from measurements to physical understanding could accelerate discovery.

Here we introduce AI Theorist (Fig.~\ref{fig:architecture}), a system of artificial intelligence (AI) agents for autonomous theoretical discovery in physics through collaboration with experimental scientists. Human scientists set the research direction and conduct experiments with assistance from AI Theorist in experimental design. From the resulting measurements, AI Theorist autonomously formulates hypotheses and iteratively designs, executes and evaluates first-principles calculations to develop physical models. These models yield physical understanding and predictions that guide further experiments. We demonstrate that AI Theorist can autonomously develop a material-specific physical model to explain previously unpublished experimental observations.

We applied AI Theorist to study the electronic structure of $\alpha$-RuCl$_3$, a leading candidate for realizing Kitaev quantum-spin-liquid physics~\cite{takagi2019kitaev,banerjee2016proximate}. Historically, advances in the microscopic understanding of crystals have helped revolutionize electronics and the semiconductor industry. The transistor, for instance, is now a foundation of modern computing~\cite{bardeen1948transistor}. This connection between fundamental understanding in materials and future technologies also motivates interest in Kitaev spin liquids, whose predicted non-Abelian excitations could support topological quantum computing~\cite{kitaev2006anyons,nayak2008nonabelian}. The electronic structure of $\alpha$-RuCl$_3$ is shaped by the interplay of electron correlations, spin–orbit coupling and crystal-field splitting, making the choice of an appropriate theoretical description particularly important~\cite{plumb2014rucl3,sandilands2016spin}. To establish a better understanding of its excitonic states, spin interactions and band structure, optical and optoelectronic measurements provide probes of this coupled physics~\cite{sandilands2016optical,binotto1971optical}.

Human scientists probed the electronic and excitonic structure of $\alpha$-RuCl$_3$ using reflection and photocurrent measurements, revealing multiple optical resonances with different prominence across the probes. Previous studies have established the importance of electronic correlations and excitonic effects in the optical response of $\alpha$-RuCl$_3$~\cite{plumb2014rucl3,sandilands2016spin,li2025quasiparticle}, while a unified first-principles description that accounts for the full set of experimentally observed optical features remains an open challenge. AI Theorist develops a more complete physical model for these previously unpublished observations through calculations of spin–orbit coupling, Coulomb screening and electron–hole interactions. The calculations identify nearby excitonic states with contrasting in-plane polarizations, providing candidate assignments for the low-energy features. Subsequent analysis by an independent human theorist corroborates the numerical methods and microscopic interpretations. Polarization-resolved predictions provide further experimental tests. Together, these results establish three contributions: autonomous discovery of physical models by an AI system from unpublished measurements; a physical explanation of the optical and photocurrent features of $\alpha$-RuCl$_3$; and a first-principles framework for autonomous theoretical discovery across materials science, connecting experimental observations to physical models and testable predictions.

\FloatBarrier
\section{AI Theorist architecture}

AI Theorist coordinates AI agents to connect experimental planning with autonomous discovery of physical models (Fig.~\ref{fig:architecture}a,b). Human scientists specify a broad research direction, in our case, understanding the electronic and excitonic structure of $\alpha$-RuCl$_3$, and AI Theorist draws on the literature to help formulate experimental questions and design appropriate measurements. After human scientists perform the experiments, AI Theorist autonomously analyses the resulting data and develops physical models to explain the observations through literature search, hypothesis generation, calculation design, execution, critique and synthesis, using literature resources and computational tools for first-principles electronic-structure and many-body calculations. Each candidate explanation is represented as an executable physical model that specifies its interactions, approximations and connection to the measured observables.

The autonomous theoretical investigation proceeds through an iterative loop of hypothesis generation, calculation design, execution and critique, followed by evidence synthesis. The hypothesis agent analyses the observations alongside relevant literature, proposes competing physical explanations and revises them in response to computational results and critique. The calculation-design agent, trained on verified computational research episodes, proposes calculations that could distinguish these explanations or resolve uncertainties in their predictions. The execution agent performs the proposed calculations using a domain-specific toolset, including Quantum ESPRESSO~\cite{giannozzi2017advanced} and BerkeleyGW~\cite{deslippe2012berkeleygw} for first-principles calculations and Bayesian optimization for parameter search~\cite{srinivas2010gaussian,yu2020hubbardbo}, and the critique agent checks their numerical reliability and quantitatively compares their predictions across experimental probes, accounting for experimental and numerical uncertainty to distinguish unreliable calculations from evidence against a physical mechanism. The results guide repairs to calculations or revisions to models, while explanations that the evidence cannot distinguish remain under consideration. The synthesis agent combines the supported results into a physical interpretation, states its limits of validity and derives predictions that inform the design of further experiments.

\FloatBarrier
\begin{figure}[!tp]
\centering
\includegraphics[width=\textwidth]{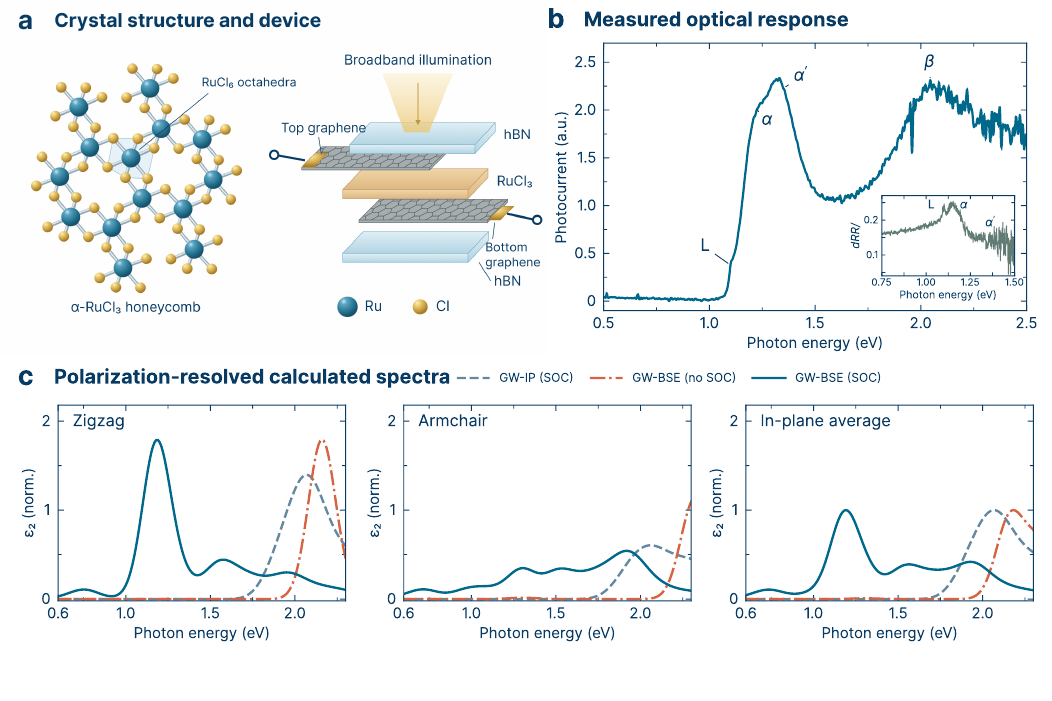}
\caption{\textbf{Crystal structure, device geometry and optical response of $\alpha$-RuCl$_3$.} \textbf{a,} Atomic structure and schematic RuCl$_3$ vertical junction device under illumination. The upper and lower graphene electrodes are contacted from opposite sides; atomic radii and device dimensions are illustrative. \textbf{b,} Photocurrent spectrum; inset shows reflection contrast, $dR/R$, over 0.75--1.50\,eV. Labels identify the experimental $L$, $\alpha$, $\alpha'$ and $\beta$ features. \textbf{c,} Zigzag, armchair and in-plane-averaged spectra from the electronic-structure and excitonic calculations: $GW$-IP (SOC), $GW$--BSE (no SOC) and $GW$--BSE (SOC). The SOC spectra use 64-point calculations; the no-SOC reference uses 36 points and is not a matched-grid SOC control. Each calculation uses one normalization factor, the maximum in-plane average over 0.6--2.65\,eV, shared across its three traces; Gaussian broadening is 0.08\,eV. The displayed window is 0.6--2.3\,eV; energies are unshifted and no additional smoothing is applied. Experimental photocurrent, reflection contrast and calculated dielectric response are distinct observables.}\label{fig:investigation}
\end{figure}

\section{AI Theorist for quantum materials}

In quantum materials, electronic-structure calculations connect the atomic arrangement of a material to its measurable properties, providing a quantitative basis for interpreting experiments. Density functional theory (DFT) offers a practical starting point by describing the ground-state distribution of electrons and the associated electronic states. Interpreting optical spectra further requires a description of the excited electrons and the holes they leave behind, including their mutual interactions. The GW approximation supplies quasiparticle energies~\cite{hedin1965new}, while the Bethe–Salpeter equation (BSE) describes how electron–hole interactions modify excitation energies and optical couplings~\cite{onida2002electronic,rohlfing2000electronhole}. These methods have become central to modern materials science, guiding the discovery and design of materials for electronic and energy technologies~\cite{jones2015density,onida2002electronic,jain2016computational}. AI Theorist uses these numerical tools to investigate how different physical hypotheses shape the optical response of $\alpha$-RuCl$_3$.

Here human scientists performed differential-reflection and photocurrent spectroscopy on a thin $\alpha$-RuCl$_3$ crystal sandwiched between graphene electrodes and encapsulated in hexagonal boron nitride (Fig.~\ref{fig:investigation}a). The differential-reflection and photocurrent measurements were carried out at 8 K, with an applied bias assisting charge collection in the photocurrent measurement. The spectra reveal three low-energy peaks, labelled $L$, $\alpha$ and $\alpha'$, together with a higher-energy feature $\beta$ (Fig.~\ref{fig:investigation}b). The $\alpha$ resonance dominates the low-energy differential-reflection spectrum and is also prominent in photocurrent. By contrast, $\alpha'$ is weak in differential reflection but more clearly resolved in photocurrent, while $L$ appears on the low-energy side of $\alpha$.

The investigation explores several computational routes before arriving at the candidate interpretation. The 64-point spinor calculations show a pronounced excitonic reconstruction of the spin–orbit-coupled optical spectrum (Fig.~\ref{fig:investigation}c). Without spin–orbit coupling (SOC), the calculated response remains weak across the experimental low-energy region, even when electron–hole interactions are included, with strong absorption developing only towards 2 eV. Including SOC at the independent-particle level still leaves little optical weight in this region. When electron–hole interactions are incorporated into the SOC calculation, substantial low-energy spectral weight emerges, with a dominant resonance near 1.19 eV. This resonance couples strongly to zigzag-polarized light and is much weaker in the armchair response. From these results and the separate matched-grid analysis, AI Theorist discovered that spin–orbit coupling and electron–hole interactions are key ingredients in the low-energy optical response of $\alpha$-RuCl$_3$.






\FloatBarrier
\section{Microscopic origin and optical selection of excitonic effects}

\begin{figure}[!t]
\centering
\includegraphics[width=\textwidth]{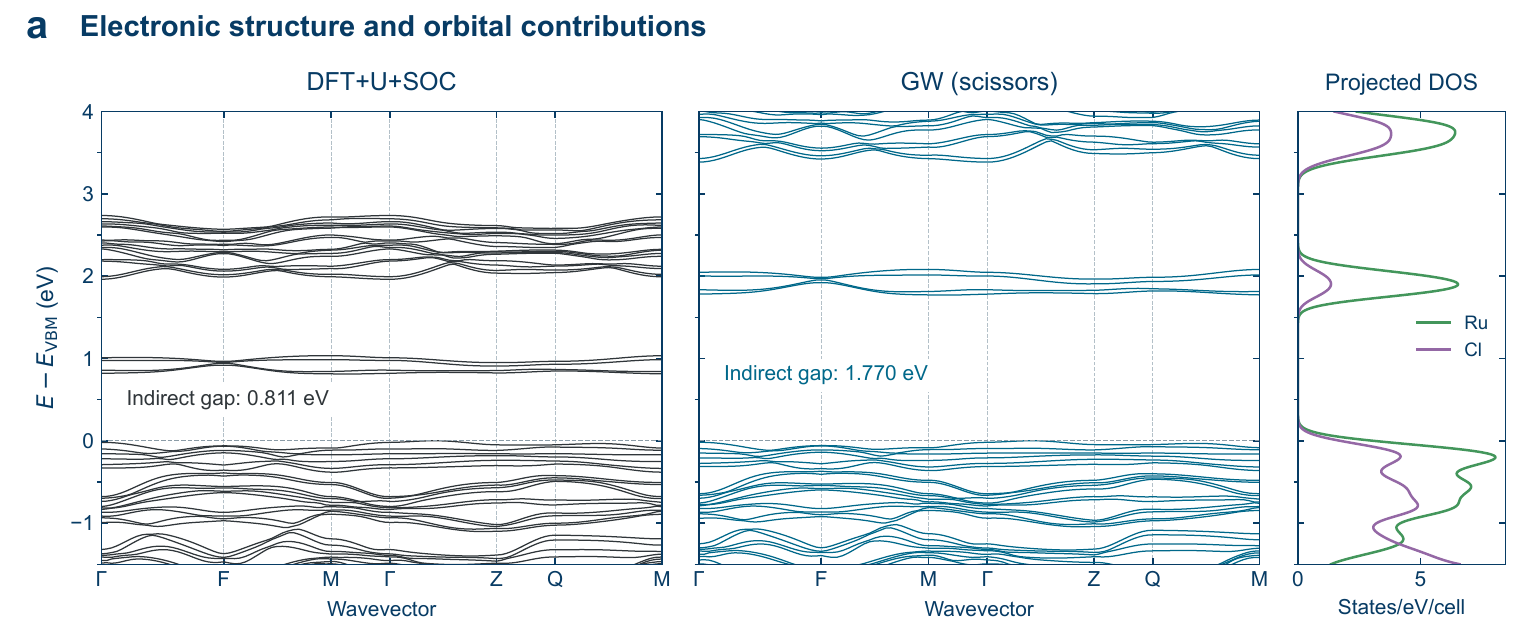}\\[2mm]
\includegraphics[width=\textwidth]{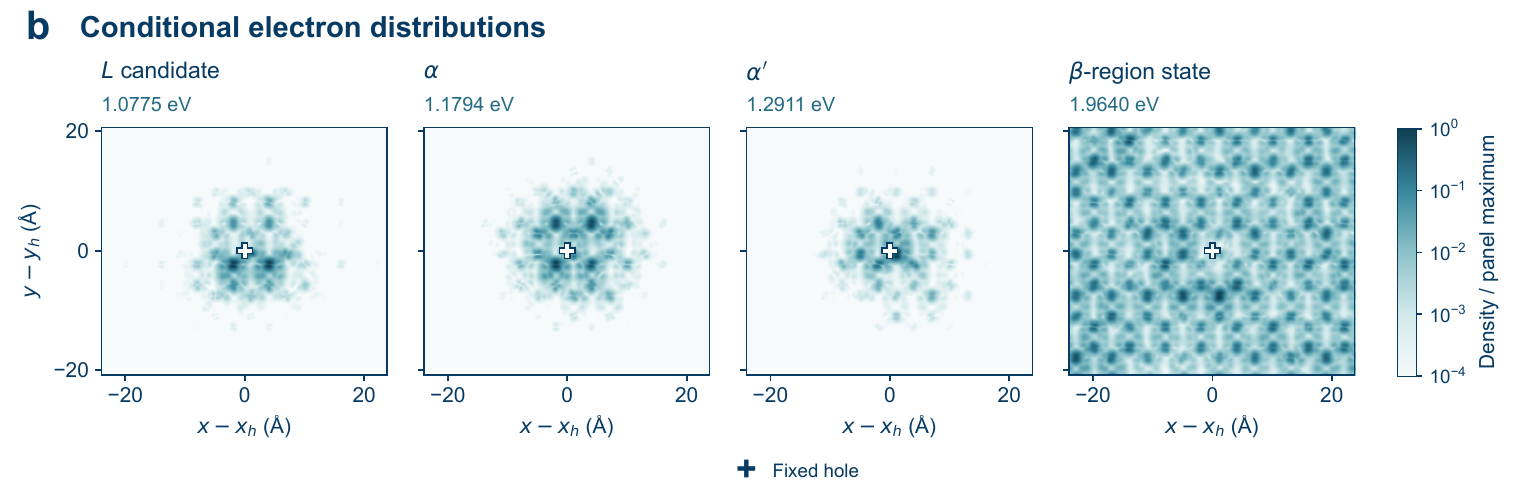}
\caption{\textbf{Electronic structure and conditional exciton distributions.} \textbf{a,} DFT+$U$+SOC bands for the 64-point spinor calculation, quasiparticle bands in the fixed two-branch $GW$ scissors representation used for the optical calculation, and Ru/Cl projected density of states. The path $\Gamma$--F--M--$\Gamma$--Z--Q--M contains 109 sampled points and 168 spinor bands. Each band panel uses its own valence-band maximum as zero; the indirect gaps along this path are 0.811 and 1.770\,eV. The $GW$ curves apply the near-gap linear corrections to the mean-field bands; they are not an explicit momentum-dependent self-energy calculation along the path. The projected DOS uses 16-point orbital projections at the corrected energies and Gaussian broadening of 0.10\,eV. \textbf{b,} Conditional electron distributions for the 64-point SOC $L$, $\alpha$ and $\alpha'$ candidates and a representative $\beta$-region state, evaluated relative to a fixed hole position. All maps use the same electron--hole spinor component $(1,1)$ and the commensurate $8\times4\times2$ periodic box. Densities are summed along the third lattice-grid direction and plotted on hole-relative in-plane coordinates; white crosses mark the fixed hole. Each map is normalized to its own maximum, with a shared logarithmic colour scale from $10^{-4}$ to 1. The comparison shows spatial distributions rather than relative integrated weights.}\label{fig:states}
\end{figure}

Figure~\ref{fig:states}a shows the calculated band structure of $\alpha$-RuCl$_3$. Applying the fixed quasiparticle correction to its DFT+$U$+SOC bands increases the indirect gap along the displayed path from 0.811 to 1.770\,eV. Both the valence- and conduction-band edges exhibit relatively flat dispersions. The projected density of states shows that the states near both band edges are dominated by contributions from Ru atoms, with smaller contributions from Cl atoms. These weakly dispersive band-edge states form the electronic basis for the low-energy electron–hole excitations.

The BSE calculation performed by AI Theorist reveals a low-energy excitonic manifold whose energy ordering and optical-strength hierarchy are consistent with the low-energy spectral features observed in Fig.~\ref{fig:investigation}b. Representative states at 1.077, 1.179 and 1.291\,eV are assigned to the experimentally labelled $L$, $\alpha$ and $\alpha'$ features, respectively. These excitation energies lie well below the approximately 1.8\,eV quasiparticle reference gap. The $\alpha$ and $\alpha'$ states are separated by approximately 0.112\,eV, with the weaker optical coupling of $\alpha'$ consistent with its reduced prominence in differential reflection.

AI Theorist further explores the conditional electron distributions to reveal the excitonic character (Fig.~\ref{fig:states}b). With a common fixed hole position and the same electron--hole spinor component, the $L$, $\alpha$ and $\alpha'$ distributions are concentrated around the hole. By comparison, the representative 1.964\,eV state in the $\beta$ region extends across much of the displayed supercell. The low-energy candidates thus share a compact conditional spatial structure, contrasting with the more extended higher-energy state. Together, their energies below the quasiparticle reference gap and their compact conditional electron distributions support an interpretation of the low-energy states as bound excitons within the adopted model.

By analysing the optical transition amplitudes, AI Theorist identifies the decisive distinction between $\alpha$ and $\alpha'$. An exciton couples to light through the coherent sum of its constituent electron--hole transitions, weighted by their polarization-dependent matrix elements~\cite{rohlfing2000electronhole}. The $\alpha$ candidate couples almost exclusively to zigzag-polarized light, with an armchair-to-zigzag squared-amplitude ratio of approximately $9\times10^{-5}$. The $\alpha'$ candidate instead favours armchair polarization, with a ratio of approximately 2.9. Summed over both in-plane polarizations, the squared transition amplitude of $\alpha'$ is approximately 5\% of that of $\alpha$. The two states consequently combine similar spatial confinement with markedly different optical strengths and preferred polarizations. Their contrasting polarization responses persist under the tested refinement of the kernel and screening grids.

The weak response below $\alpha$ provides a further signature of the relativistic excitonic structure. A comparison of spinor and scalar-relativistic calculations on matched 36-point kernel and screening grids reveals a pronounced redistribution of discrete optical weight. The satellite-to-reference ratio of summed squared transition amplitudes is approximately 0.093 in the SOC calculation and $3\times10^{-5}$ in the scalar-relativistic reference. This contrast precedes spectral broadening and supports an SOC-related interpretation of the weak low-energy response associated with $L$. Alongside the distinct optical couplings of $\alpha$ and $\alpha'$, it reveals substantial variation in optical activity within the low-energy exciton manifold.

\begingroup\interlinepenalty=10000
AI Theorist uses this hierarchy of optical couplings to interpret the different prominence of the resonances in reflection and photocurrent. With the common 0.08\,eV Gaussian broadening, the weak $L$ and $\alpha'$ contributions overlap with stronger neighbouring transitions in the polarization-averaged dielectric response. Photocurrent additionally weights each excitation by its probability of producing collected charge, which could introduce a weight reconstruction among different excitations~\cite{massicotte2018dissociation}. Relative to a common electron–hole continuum, the higher excitation energy of $\alpha'$ implies weaker binding, which could facilitate field-assisted dissociation. This can potentially account for the difference in visibility of $\alpha'$ in reflection and photocurrent spectra. State-dependent dissociation and collection therefore provide a route for resolving an optically weak exciton in the electrical response. A future polarization-resolved spectroscopy study would further explore their contrasting optical responses, particularly the predicted enhancement of $\alpha'$ relative to $\alpha$ as the incident polarization rotates from zigzag to armchair.
\par\endgroup

\FloatBarrier
\begin{figure}[!t]
\centering
\includegraphics[width=\textwidth]{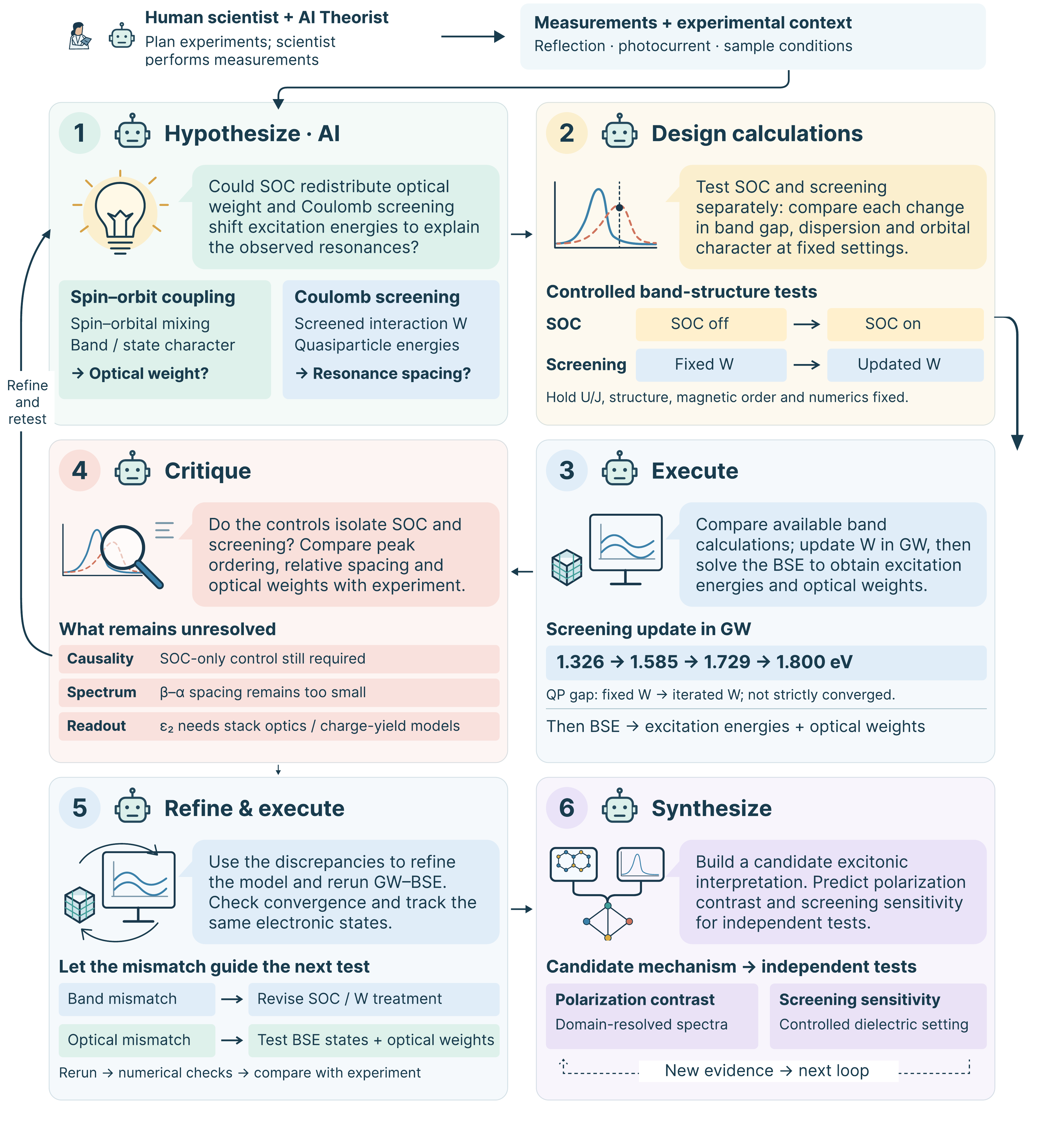}
\caption{\textbf{From experimental observations to a candidate physical interpretation.} Human scientists supply reflection and photocurrent spectra together with sample context. AI Theorist hypothesizes roles for spin--orbit coupling and Coulomb screening, designs separate tests of their effects on the band structure, and evaluates optical consequences using GW--BSE calculations. Critique addresses causal isolation, spectral discrepancies and the relation between calculated and measured observables; these checks guide model refinement and proposed polarization and screening tests.}\label{fig:interactions}
\end{figure}

\section{Analysis of the discovery process}

We investigate the optical response of $\alpha$-RuCl$_3$, motivated by its promise as a platform for Kitaev quantum-spin-liquid physics and the potential of such states for topological quantum computing. We target low-energy optical resonances, where electron–hole interactions strongly shape the excitation spectrum~\cite{sandilands2016optical,sinn2016electronic,li2025quasiparticle}. Optical and optoelectronic measurements provide complementary probes of the material's electronic structure, excitonic states, and spin interactions. Human scientists set the research direction and conducted measurements with AI Theorist assisting in experimental design; the resulting reflection and photocurrent spectra provided the starting point for autonomous model development.

AI Theorist used 153.98 million tokens, including 0.68 million output tokens, to investigate the observations through literature analysis, hypothesis generation and computational testing. AI Theorist proposed and then investigated three explanatory routes: local correlations and multiplet excitations, a constrained electronic starting state, and a model combining relaxed orbital occupations with updated screening and electron--hole interactions. The agents rejected the tested local-correlation model because it could not jointly explain the gap and optical features, and the constrained starting state because it produced an unobserved dominant optical peak, then refined the third route.

The reasoning behind this transition linked each spectral discrepancy to a physical assumption that could be tested (Fig.~\ref{fig:interactions}). Releasing the occupation constraint reduced the anomalous peak and restored $\alpha$ as the dominant low-energy resonance, directing the investigation towards the unconstrained electronic state. The agents then examined whether the remaining gap discrepancy arose from the quasiparticle energies or the screening response. Updating the energies while holding screening fixed changed the gap only slightly, motivating calculations that also updated screening. These screening updates increased the calculated gap from approximately 1.33 to 1.80 eV, providing the electronic starting point for subsequent excitonic calculations.

The resulting model describes the low-energy optical features as bound electron--hole excitations with distinct optical selection rules. An exciton's coupling to light depends on how its constituent electronic transitions combine for each polarization. AI Theorist identified nearby $\alpha$ and $\alpha'$ candidates that preferentially couple to zigzag and armchair polarization, respectively. The weaker total optical coupling of $\alpha'$ accounts for its reduced prominence in the calculated polarization-averaged response. Because photocurrent also depends on exciton dissociation and charge collection, this distinction provides a possible explanation for the different visibility of $\alpha'$ in reflection and photocurrent.

AI Theorist tested the interpretation through refinements of the excitonic calculations and analyses of the states' spatial structure. These checks address complementary questions: whether the optical assignments persist as the numerical treatment changes and whether the states have the character expected of bound excitons. The contrasting polarizations of $\alpha$ and $\alpha'$ persisted under the tested grid refinements. Their compact electron--hole distributions and energies below the quasiparticle gap further supported the excitonic interpretation (Fig.~\ref{fig:states}). Subsequent analysis by an independent human theorist corroborated the numerical methods and microscopic interpretations. The investigation thus produced a microscopic model and a discriminating prediction: rotating the incident polarization from zigzag to armchair should enhance $\alpha'$ relative to $\alpha$, providing a test for future measurements.

\FloatBarrier
\section{Discussion}

This study demonstrates autonomous development of a microscopic physical model to explain previously unpublished optical and photocurrent observations in a quantum material. Human scientists set the research direction and performed the measurements with assistance from AI Theorist in experimental design, after which the system developed and tested candidate explanations. The investigation required connecting electronic correlations and electron--hole interactions to optical selection rules and the different responses of the experimental probes. The calculations identify nearby excitonic states with contrasting polarization responses, providing candidate assignments for the low-energy features of $\alpha$-RuCl$_3$. Our calculations identify spin--orbit coupling, Coulomb screening, and electron--hole interactions as key ingredients for explaining the observed low-energy optical features of $\alpha$-RuCl$_3$.

Previous work has demonstrated equation discovery, automated research workflows and AI-generated scientific hypotheses~\cite{schmidt2009laws,udrescu2020aifeynman,cornelio2023descartes,swanson2025virtual,gottweis2026coscientist,ghareeb2026robin}. By contrast, AI Theorist uses first-principles calculations to autonomously develop and test microscopic physical models that explain previously unpublished experimental observations. These calculations connect hypotheses about microscopic interactions to predicted excitation energies, optical strengths and polarization dependence, allowing competing explanations to be assessed against the measurements. With human scientists setting the overall scientific direction and providing measurements, the agents select computational tests, assess discrepancies and revise the physical interpretation (Fig.~\ref{fig:interactions}). The distinction between calculated excitonic states and resolved spectral peaks is central to this assessment: candidate assignments must account for optical selection rules and the response of each experimental probe, rather than energy agreement alone.

\begin{samepage}
Beyond explaining the measured spectra, the proposed excitonic picture opens a route to exploring how optical excitations couple to magnetism in $\alpha$-RuCl$_3$~\cite{sandilands2016optical}. Combined with temperature-dependent measurements and magnetic field dependence, these studies could test whether the proposed excitonic states provide optical probes of magnetic order and spin correlations. Another open question is how charge transfer and screening at graphene contacts modify this response in the two-dimensional limit~\cite{biswas2019proximity,rossi2023charge}.
\par
\end{samepage}

AI Theorist points towards a mode of discovery in which the construction and testing of microscopic theory can proceed autonomously alongside experimental measurement. First-principles calculations give proposed mechanisms quantitative consequences, enabling agents to revise physical models against evidence and derive predictions beyond the observations that motivated them. Applied across materials and measurement techniques, such systems could compare competing interactions, identify mechanisms underlying unexpected responses and propose measurements that distinguish the remaining explanations. This capability could help turn growing experimental datasets into physical understanding of how structure, composition and external fields govern material properties. Extending the approach will require scientific tools and validation suited to each problem, with human scientists setting research priorities and conducting the experiments. We envision a future in which AI Theorist works alongside human scientists and automated physical laboratories to accelerate discovery in materials science. By connecting autonomous model discovery to experimentally testable predictions, AI Theorist offers a path towards uncovering microscopic mechanisms and using that understanding to guide the search for materials with desired properties.

\FloatBarrier

\clearpage

\section{Methods}

\subsection{AI Theorist architecture and implementation}
AI Theorist is a system of AI agents for autonomous discovery of physical models from experimental observations (Fig.~\ref{fig:architecture}). Its hypothesis, calculation-design, execution, critique and synthesis agents carry out the functions shown in Fig.~\ref{fig:architecture}a, with literature search supporting hypothesis formation and experimental planning linking the system to human measurements. The hypothesis agent ranks candidate explanations using the literature and available evidence, the calculation-design agent specifies tests of those explanations, and the execution agent performs the calculations using its scientific tools. Critique determines whether the results call for numerical repair, model revision or further investigation, while synthesis combines the supported results into a physical interpretation and testable predictions. The implementation uses Claude Opus 4.6~\cite{anthropic2026opus46} for the hypothesis, execution, critique and synthesis agents, and Qwen3.5-27B~\cite{qwenteam2026qwen35} adapted through supervised fine-tuning (SFT)~\cite{ouyang2022training} and group relative policy optimization (GRPO)~\cite{shao2024deepseekmath} for the calculation-design agent. Role-specific prompts define each agent's responsibilities, while scientific tools support execution and a critique rubric guides assessment. Role prompts, tool permissions and configuration details will be shared with the code release once available.

\subsection{Human--AI research workflow}
Human scientists initiate an investigation by specifying a broad research direction, such as understanding the electronic and excitonic structure of $\alpha$-RuCl$_3$. AI Theorist uses the literature to help formulate experimental questions and propose measurements. In the RuCl$_3$ case study, differential-reflection and photocurrent measurements provide complementary probes of optical excitations and their conversion into collected charge. Human scientists carry out the experiments and supply the resulting measurements, acquisition conditions and relevant sample information. AI Theorist then investigates the observations through the hypothesis--calculation-execution--critique loop, developing physical models and deriving testable predictions. Human contributions include setting the research direction, conducting the measurements and providing experimental context.

\subsection{AI Theorist toolset}
The tool registry exposes each capability as a named callable module with defined inputs, returned quantities or evidence, and usage constraints. The execution agent checks these inputs before running a calculation and uses failure diagnostics to request bounded repairs. For literature-dependent questions, the hypothesis agent uses Paperclip~\cite{gxl2026paperclip} to locate papers and inspect passages relevant to a candidate explanation. Source references, calculation inputs, outputs and diagnostics are retained in the investigation record for critique and synthesis.

The registry comprises 18 tools spanning literature retrieval, electronic-structure and many-body calculations, and numerical analysis. The tools comprise four Paperclip~\cite{gxl2026paperclip} interfaces for literature search, paper preview, paper or metadata reading and passage search; three Quantum ESPRESSO~\cite{giannozzi2017advanced} interfaces for electronic-structure calculations, orbital projections and export to BerkeleyGW; seven BerkeleyGW~\cite{deslippe2012berkeleygw} interfaces for wavefunction-format conversion, dielectric screening, quasiparticle corrections, electron--hole kernels, excitonic spectra, quasiparticle interpolation and real-space exciton analysis; a GP-UCB Python module for parameter search; a numerical-analysis Python module using NumPy~\cite{harris2020array} and SciPy~\cite{virtanen2020scipy}; a model-fitting Python module using SciPy optimization routines; and a visualization Python module using Matplotlib~\cite{hunter2007matplotlib}. The hypothesis agent selects literature tools, while the calculation-design agent specifies scientific tasks and the execution agent invokes the corresponding computational modules.

Literature queries use the material, observable and physical mechanism under investigation to identify relevant papers. The hypothesis agent then inspects paper text and metadata, retaining source identifiers and relevant passages rather than relying on search summaries alone. This retrieval process provides traceable evidence for hypothesis formation and model comparison.

Calculations used Quantum ESPRESSO 7.4.1 and the complex build of BerkeleyGW 4.0, with NumPy 2.4.6 and h5py 3.16.0 for numerical analysis and data handling.

The parameter-search module uses GP-UCB with physical bounds and the comparison objective supplied by the calculation-design agent.

\subsection{Calculation-design agent training and evaluation}
Computational investigations require sequences of calculation decisions, in which an unsuitable approximation or an unresolved numerical error can undermine the final physical interpretation. To address this problem, we trained the calculation-design agent to select computations using rewards based on the scientific conclusions supported by their executed results.

After supervised fine-tuning of Qwen3.5-27B, GRPO optimized its calculation-selection policy in a bounded environment of small electron--hole and Hubbard calculations. The objective accepts different calculation sequences that establish the correct supported model set or a justified unresolved conclusion, without requiring one prescribed sequence of actions. During training, the agent generated four rollouts per task and received a terminal reward determined by numerical accuracy and evidential support, with a computation-cost penalty applied only to successful episodes.

Group-relative advantages derived from these terminal rewards were used to update the planner decisions across each rollout while the executor and numerical scorer remained fixed. The training objective encourages informative calculation selection, repair of insufficient numerical resolution and discrimination between numerical failure and evidence against a physical model. Evaluation assessed multistep calculation reliability, while the outcome-based training procedure avoided requiring a human annotation for every intermediate decision. The training corpus comprised public MatRL tool-use episodes~\cite{amayuelas2026matrl} and execution-checked physics trajectories for SFT, followed by additional GRPO tasks covering model discrimination, numerical reliability and parameter estimation. We compared the original, SFT-only and SFT+GRPO Qwen3.5-27B policies with Claude Opus 4.6 on prespecified synthetic tasks and Quantum ESPRESSO transfer cases under matched tools and budgets, using separate validation tasks for development and excluding all RuCl$_3$ case-study observations and conclusions from training and checkpoint selection.


\subsection{Gaussian-process upper-confidence-bound optimization for parameter search}
Scientific question-answering performance need not translate into successful experimental reasoning, as shown in ScienceWorld~\cite{wang2022scienceworld}. In our setting, the calculation-design agent must select physical-model parameter values whose calculated observables reduce disagreement with the measurements, while limiting the number of expensive electronic-structure and many-body calculations. A brute-force parameter sweep can consume this budget without using previous results to prioritize the next evaluation. To address this problem, we integrate a GP-UCB parameter-search module~\cite{srinivas2010gaussian}, drawing on the DFT+$U$ optimization approach of Yu et al., who used a Gaussian-process surrogate and UCB acquisition to tune Hubbard parameters against HSE reference band gaps and band structures \cite{yu2020hubbardbo}.

The calculation-design agent specifies the physical model, adjustable parameters, admissible bounds and comparison objective, and the execution agent invokes GP-UCB to select successive evaluations using the results of previous evaluations. A Gaussian-process surrogate fitted to completed calculations estimates the objective mean and uncertainty across the search domain, and the acquisition function balances predicted agreement with exploration of uncertain parameter regions.

For a discrepancy $L(\theta)$ to be minimized, we model $f(\theta)=-L(\theta)$ and select the next parameter vector according to

\begin{equation}
\theta_{t+1}=\underset{\theta\in\Theta}{\operatorname{arg\,max}}\left[\mu_t(\theta)+\kappa_t\sigma_t(\theta)\right],
\end{equation}

where $\Theta$ is the admissible parameter domain, $\mu_t$ and $\sigma_t$ are the posterior mean and standard deviation of $f$ after $t$ evaluations, and $\kappa_t$ controls exploration. The execution agent runs the selected calculation, records its numerical diagnostics and updates the surrogate with valid results, while failed calculations return diagnostic feedback without being counted as improved agreement. The search terminates at its specified stopping criterion or budget, and the best valid result is passed to critique together with its calculation record. GP-UCB searches parameters within the specified physical model; hypothesis generation and revision remain the responsibility of the hypothesis agent.

\subsection{\texorpdfstring{Application to $\alpha$-RuCl$_3$}{Application to α-RuCl3}}
The scientific objective was to develop a physical model of the electronic and excitonic structure of $\alpha$-RuCl$_3$ that could explain the optical and photocurrent observations. The investigation used measurements and sample context supplied by human scientists, together with literature on the material. AI Theorist investigated spin-orbit coupling, quasiparticle screening and electron--hole interactions through successive calculations, with the principal numerical results obtained from the 64-point spinor calculation described below.

\subsection{Electronic-structure and many-body calculations}
Electronic-structure and many-body calculations were used to test how quasiparticle corrections, electron--hole interactions and spin--orbit coupling affect the predicted excitation spectrum. The electronic-structure calculations used a 16-atom monoclinic bulk structure with zigzag antiferromagnetic order~\cite{johnson2015monoclinic}, Quantum ESPRESSO 7.4.1~\cite{giannozzi2017advanced} and the complex build of BerkeleyGW 4.0~\cite{deslippe2012berkeleygw}. The spinor starting point used PBE~\cite{perdew1996pbe} with the Liechtenstein DFT+$U$+$J$ formulation~\cite{liechtenstein1995density} applied to the Ru $4d$ orbitals using atomic projectors, with $U=3.6$ eV and $J=0.4$ eV, followed by GW quasiparticle corrections~\cite{hedin1965new} and a Bethe--Salpeter calculation in the Tamm--Dancoff approximation~\cite{onida2002electronic,rohlfing2000electronhole}. Calculated observables comprise the imaginary dielectric response along the two in-plane directions, excitation energies, optical matrix elements and conditional electron distributions. The conditional-density maps use spinor component $(1,1)$, with each map normalized to its own maximum.

The screening calculation used 440-band sums and three-dimensional periodic screening without Coulomb truncation. The GW self-energy was evaluated using the Hybertsen--Louie generalized plasmon-pole approximation~\cite{hybertsen1986electron}, with a static-remainder correction to the Coulomb-hole term~\cite{deslippe2013remainder}. Screening updates yielded a final calculated gap of 1.8004 eV and an extrapolated limiting value of 1.8710 eV. The 64-point excitonic calculation retained the near-gap linear quasiparticle correction from the last completed screening update. Grid sensitivity was assessed by comparison with the preceding 36-point kernel calculation and by 64-to-96-point spectral interpolation. The $\alpha$ and $\alpha'$ candidates lie at 1.179386 and 1.291078 eV, with changes of $-$13.2 and $-$12.2 meV from 36 to 64 points, respectively. The dominant-channel centroid changes by at most 8.3 meV under 64-to-96-point interpolation. These comparisons quantify grid sensitivity at the retained quasiparticle correction, without establishing convergence of the mean-field or screening calculations.

Spectral assignments were examined using discrete excitation energies, polarization-resolved optical strengths and the broadened dielectric response. At the adopted broadening, the $L$ and $\alpha'$ contributions do not form distinct maxima in the unshifted, in-plane-averaged spectrum. The SOC spectra in Fig.~\ref{fig:investigation}c use a common 64-point grid, whereas the scalar-relativistic reference uses 36 points; the comparison between these calculations therefore does not isolate SOC alone. The scalar-relativistic and spinor results compare electronic-structure treatments that differ in pseudopotentials and quasiparticle corrections as well as spin--orbit coupling; they therefore characterize the combined change in electronic structure.

\subsection{Experimental measurements and spectral analysis}
$\alpha$-RuCl$_3$, graphene/graphite (HQ Graphene) and hBN (HQ Graphene) flakes were mechanically exfoliated onto silicon dioxide substrates in a nitrogen-filled glovebox, with oxygen and water concentrations below 0.1~ppm. Sapphire substrates were patterned with metal contacts (3~nm Ti/17~nm Au) and wire-bonding pads (5~nm Ti/40~nm Au) using photolithography and electron-beam deposition. The heterostructures were assembled inside the glovebox by sequential dry transfer with a polycarbonate-coated polydimethylsiloxane stamp. A thin $\alpha$-RuCl$_3$ flake was placed between two graphene electrodes and encapsulated in hBN to form a vertical junction. The completed devices were wire-bonded to chip carriers for measurements.

Photocurrent and differential-reflection measurements were conducted under vacuum at 8~K in a Montana Instruments optical cryostat. Both measurements used a Bruker IFS 66v/S Fourier-transform infrared (FTIR) spectrometer with a tungsten--halogen source.

For photocurrent spectroscopy, the illumination passed through the FTIR interferometer and was focused onto the device using a parabolic mirror. A d.c.\ bias was applied between the graphene electrodes using a Keithley 2450 source measure unit. The photocurrent signal was converted to a voltage using a Stanford Research Systems SR570 current preamplifier, further amplified using an SR560 voltage preamplifier, and fed into the FTIR acquisition system. Fourier transformation of the recorded interferogram yielded the photocurrent spectrum at each applied bias. The photocurrent spectra were divided by the corresponding source spectra measured using InGaAs and silicon reference detectors to account for the spectral distribution of the illumination.

For differential-reflection spectroscopy, the illumination from the FTIR interferometer was focused onto the sample using a reflective objective with a numerical aperture of 0.5. The reflected light was collected with an InGaAs detector. The reflection contrast was calculated as
\[
\frac{\Delta R}{R}
=
\frac{R_{\mathrm{sample}}-R_{\mathrm{ref}}}{R_{\mathrm{ref}}},
\]
where $R_{\mathrm{sample}}$ and $R_{\mathrm{ref}}$ denote the sample and reference reflection spectra, respectively. Details of the reference measurement will accompany the data release.

\subsection{Computational resources and token usage}
Total model usage was 153.30 million input tokens and 0.68 million output tokens across the two models, for 153.98 million tokens in total. Qwen3.5-27B accounts for approximately 57\% of input tokens and 60\% of output tokens; Claude Opus 4.6 accounts for the remaining 43\% and 40\%, respectively. Percentages are rounded within each column, and token counts follow each model's tokenizer.

\noindent\begin{minipage}{\textwidth}\small\setlength{\tabcolsep}{4pt}\renewcommand{\arraystretch}{1.15}
\begin{tabular}{@{}p{0.28\textwidth}p{0.15\textwidth}p{0.14\textwidth}p{0.15\textwidth}p{0.14\textwidth}@{}}
\toprule
\textbf{Model} & \textbf{Input tokens} & \textbf{Share of input} & \textbf{Output tokens} & \textbf{Share of output} \\
\midrule
Qwen3.5-27B & 87.70 M & 57\% & 0.41 M & 60\% \\
Claude Opus 4.6 & 65.60 M & 43\% & 0.27 M & 40\% \\
\textbf{Total} & \textbf{153.30 M} & \textbf{100\%} & \textbf{0.68 M} & \textbf{100\%} \\
\bottomrule\end{tabular}\end{minipage}\par\medskip

Selected scientific calculations took 2.5 h for the fixed-screening calculation, 38.5 h for the screening-update sequence and 27.9 h for the subsequent 36-point calculation. The first duration excludes an earlier attempt that waited in the queue without running, the second includes an interruption and restart, and the third has separately recorded waiting and kernel-computation times of 1.4 h and 2.3 h. Because these timing boundaries differ, the durations are reported separately. The 36-point calculation used one node with 16 MPI processes, four momentum-point pools for the plane-wave calculation and one numerical-library thread per process.

Qwen3.5-27B inference used vLLM~\cite{kwon2023pagedattention} on one NVIDIA H200 GPU. For Claude Opus 4.6, the standard prices announced on 5 February 2026 were US\$5 per million input tokens and US\$25 per million output tokens~\cite{anthropic2026opus46}. Applying these rates gives an estimated token cost of $65.60\times5+0.27\times25=\mathrm{US}\$334.75$.

\bmhead{Data availability}
The data supporting this study, including experimental and numerical records, calculation tables, analysis scripts and figure source data, will be shared in a public repository once available. Repository identifiers and licensing information will accompany the data release.

\bmhead{Code availability}
The agent implementation, role prompts, tool configurations, training and evaluation scripts, checkpoints, and figure-generation code will be shared once available. The repository URL, version and license will accompany the code release.

\backmatter

\end{document}